\documentclass[a4paper]{jpconf}
\usepackage{graphicx}
\usepackage{comment}
\begin{document}
\title{Global Transformer Architecture for Indoor Room Temperature Forecasting\footnote{The paper will be published in the CISBAT 2023 Special Issue of the Journal of Physics Conference Series, Vol. 2600}}

\author{Alfredo V Clemente$^1$, Alessandro Nocente$^1$, and Massimiliano Ruocco$^{1,2}$}
\address{$^1$ SINTEF AS, Trondheim, Norway}
\address{$^2$ Norwegian University of Science and Technology, Trondheim, Norway}

\ead{massimiliano.ruocco@sintef.no}

\begin{abstract}
A thorough regulation of building energy systems translates in relevant energy savings and in a better comfort for the occupants. 
Algorithms to predict the thermal state of a building on a certain time horizon with a good confidence are essential for the implementation of effective control systems.  This work presents a global Transformer architecture for indoor temperature forecasting in multi-room buildings, aiming at optimizing energy consumption and reducing greenhouse gas emissions associated with HVAC systems. Recent advancements in deep learning have enabled the development of more sophisticated forecasting models compared to traditional feedback control systems. The proposed global Transformer architecture can be trained on the entire dataset encompassing all rooms, eliminating the need for multiple room-specific models, significantly improving predictive performance, and simplifying deployment and maintenance. Notably, this study is the first to apply a Transformer architecture for indoor temperature forecasting in multi-room buildings. The proposed approach provides a novel solution to enhance the accuracy and efficiency of temperature forecasting, serving as a valuable tool to optimize energy consumption and decrease greenhouse gas emissions in the building sector.

\end{abstract}
\section{Introduction and Related Work}
According to the latest IPPC report \cite{IPCC_2022_WGIII_Ch_9} the building industry has the potential to reduce its GHG emissions by up to 66\%.
Building operation is one of the main contributor to this impact, and most of it is to be attributed to heating and cooling of residential and commercial building.

Indoor temperature forecasting plays a critical role in optimizing the performance of HVAC systems, which are responsible for a significant portion of the energy consumption and associated greenhouse gas emissions in buildings. Traditional feedback control systems based on a set-point value do not always take into account the dynamic nature of the thermal environment and can result in inefficient energy use, including overshooting the set-point and unnecessary heating or cooling.

Recent advancements in machine learning and deep learning \cite{TIEN2022100198} have enabled the development of more sophisticated indoor temperature forecasting models that can capture the complex interactions between internal and external factors that influence the thermal state of a space. These models are based on a range of inputs, including weather data, occupancy patterns, building characteristics, and HVAC system performance, and use advanced algorithms to generate accurate and reliable predictions of the thermal state of a space.

In this work, we propose a global Transformer architecture, based on the original vanilla Transformer \cite{Vaswani2017}, to forecast indoor room temperature in a multi-room building. 
The Transformer architecture offers several advantages over other statistical machine learning approaches and other deep learning based architecture, such as LSTM networks, including the ability to manage inputs of different lengths and include future covariates. Additionally, the Transformer architecture is highly parallelizable, allowing us to perform experiments on a large-scale dataset. The proposed model is trained on the total set of data over all the rooms, providing the advantage of having a single model for all rooms, as opposed to a single model per room, which can be challenging to maintain. 
To incorporate the room ID information into the unified architecture, we introduce a novel approach that avoids the need for a separate model for each room. 
To the best of our knowledge, we are the first to employ a Transformer architecture for indoor temperature forecasting in a multi-room building. It also offers several benefits over traditional approaches to indoor temperature forecasting. 
By utilizing a single model for all rooms, we can reduce the burden of maintaining multiple models and improve the overall efficiency of the forecasting process. 
\section{Methods and Dataset}
\subsection{Dataset}
We considered a dataset containing data from $133$ rooms $r \in \mathbf{R}$ in a single building, with a total of 839 time series. These are distributed as follows:
\begin{itemize}
\item 29 building sensors that are common across all rooms, such as water flows, water temperatures, solar shading, among others.
\item 5 weather forecast variables shared across all rooms such as solar radiation, relative humidity, air temperature, dew point and cloud coverage.
\item 7 variables related to the date and time, such as the day of the week and hour of the day, shared across all rooms.
\item 5 room-specific variables, such as air temperature setpoint and whether cooling was applied to the room.
\item The target variable: room air temperature.
\end{itemize}

The dataset has an hourly resolution and covers approximately two years, consisting of 19,115 hours. The data is split into train, validation, and test sets with an 82\%, 14\%, and 4\% split, respectively, in chronological order. 

The time series are categorized into two sets based on their availability at inference time. 
The target series $Y_{r_i}$ and past covariates $C^p_{r_i}$ are known only until the inference point, while the future covariates $C^f_{r_i}$ are known for the forecasting horizon and the input window. All timeseries are past covariates, while only the weather forecasts, date and time related variables and the known setpoints are future covariates.

Finally, each room is assigned with an id value of id $ 0 < i \leq |\mathbf{R}| - 1$.

\subsection{Proposed Models}
The goal is to produce a model that is able to predict the room temperature of 133 rooms in a large office building using building sensors, weather forecasts and other available data.
In this study, we compare three different types of models, namely a baseline persistence model, a multi-layer LSTM neural network, and a proposed transformer model. For each neural network, we perform a hyperparameter search to determine the best hyperparameters. 
For all the considered models (a part from the persistence model) the objective is to approximate the function
\begin{equation}
f(Y_{r_i,(t-k...t)}, C^p_{r_i,(t-k...t)}, C^f_{r_i,(t+1...t+n)})=Y_{r_i,(t+1...t+n)} 
\label{Eq:1}
\end{equation}
In Equation \ref{Eq:1}, $Y_{r_i}$ denotes the temperature of room $r_i$, $C^p_{r_i}$ represents the covariates that are known only within the range $[t-k, t]$, and $C^f_{r_i}$ is the set of covariates that are known within the range $[t-k, t+n]$, known as \textit{future covariates}.

Both neural network models are residual models relative to the persistence model, this means they predict

\begin{equation}
F(Y_{r_i,(t-k...t)}, C^p_{r_i,(t-k...t)}, C^f_{r_i,(t+1...t+n)}, i|\theta)=\bar{Y}_{r_i,(t+1...t+n)} 
\end{equation}

where
\begin{equation}
Y_{r_i,(t+1...t+n)} \approx \bar{Y}_{r_i,(t+1...t+n)} + \mathbf{1}Y_{r_i, (t)}
\label{eq:residual}
\end{equation}

Here follow more details about the model considered.\\

\textbf{Persistence model}. Given that indoor room temperatures are highly correlated in time, a simple and reasonable baseline mode is a persistence model. This model is defined as 

\begin{equation}
F(Y_{r_i,(t-k...t)}, C^p_{r_i,(t-k...t)}, C^f_{r_i,(t+1...t+n)}) = \mathbf{1}Y_{r_i, (t)}
\label{eq:persistence}
\end{equation}

meaning the model simply uses the room temperature $Y_{r_i, (t)}$ as the estimate for the room temperature for the next $n$ hours. 

\textbf{LSTM}. This neural network is based on an encoder-decoder architecture that uses LSTM \cite{hochreiter1997long} layers. The network includes 8 LSTM layers in both the encoder and decoder, with each layer having 32 units. To feed the encoder, the past covariates $C^p$ and target $Y$ are concatenated along the channels axis. The hidden state and cell state of the last encoder layer are used to initialize the first decoder layer's hidden state and cell state. The decoder layer's input consists of future covariates $C^f$. The output of the last decoder layer at each timestep is flattened and passed through a linear layer with a RELU non-linearity and a size of 256. The resulting output is then passed through another linear layer with a size of $n$ to produce the final output.

\textbf{Transformer}. The proposed model is an encoder-decoder transformer \cite{vaswani2017attention} improved using well known modern methods. The original sinusoidal positional encoding is replaced  with a rotary position encoding (RPE) \cite{su2021roformer}, the RELU activation is replaced with gated linear units (GLU) \cite{glu} activation, the LayerNorm \cite{layernorm} is replaced with ScaleNorm \cite{scalenorm}, finally the normalization layer is moved to be the first layer of a block instead of the last (PreNorm) \cite{prenorm}. These improvements were chosen as they have been shown to increase the performance of Transformers when modeling sequences \cite{su2021roformer, glu_improve, scalenorm, layernorm, prenorm}. 

Each encoder block is comprised of a ScaleNorm layer, followed by a self-attention layer of size $32$, a linear layer of size $128$ and finally a GLU activation. There are residual connections between the encoder blocks. The encoder consists of 4 encoder blocks. Similary, each decoder block is comprised of a ScaleNorm layer, followed by  a self-attention layer of size $32$, a cross-attention layer of size $32$ with the output of the encoder, a linear layer of size $128$, and finally a GLU activation. The output of the decoder is flattened and passed to a linear layer of size $n$.

\begin{figure}
\begin{center}
\includegraphics[width=0.9\textwidth]{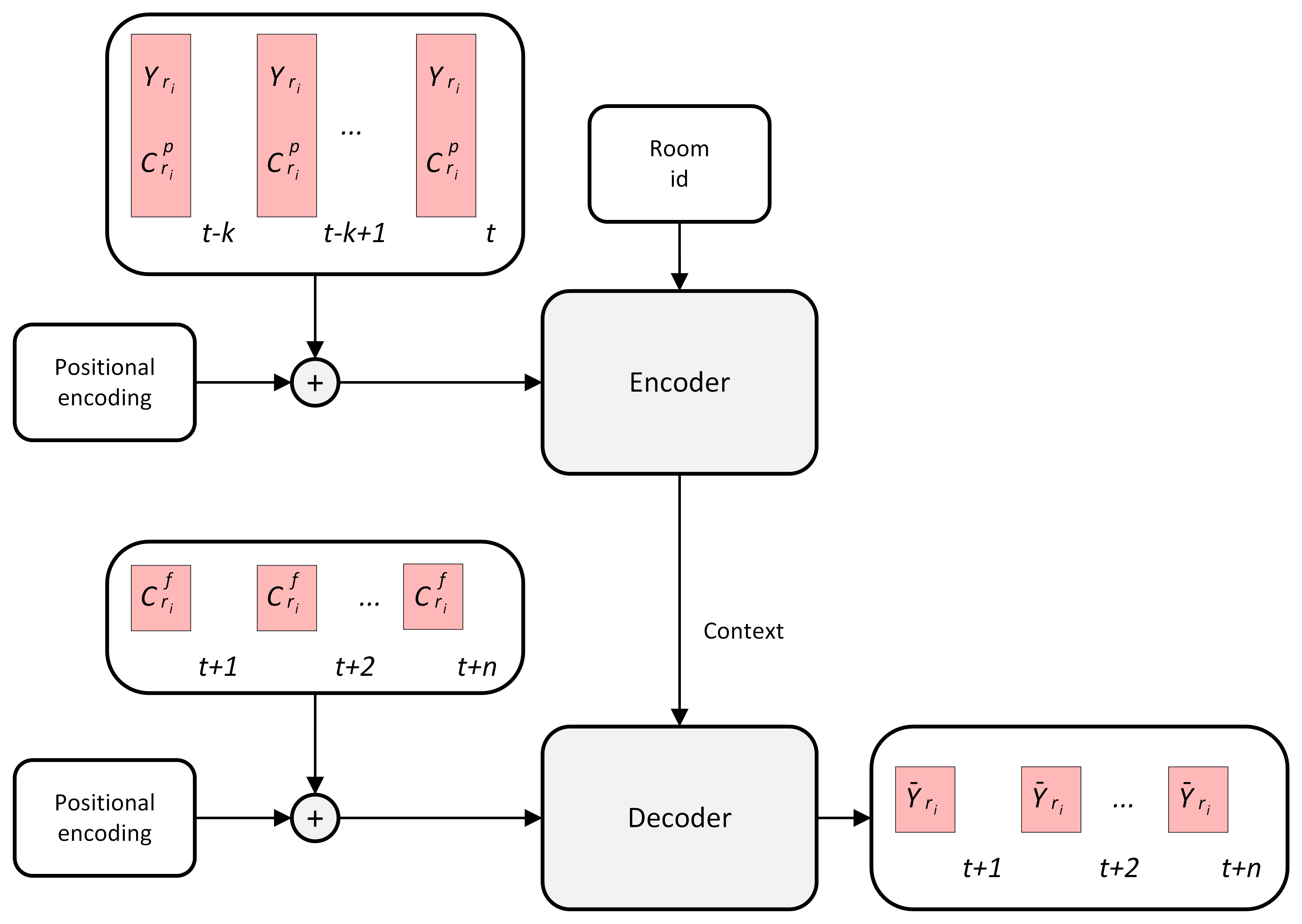}
\end{center}
\caption{\label{fig:transformer} Encoder decoder transformer model following the original paper \cite{vaswani2017attention} with the addition of the room id.}
\end{figure}
\section{Experimental Settings}
In order to fairly compare the methods, each experiment is repeated 8 times with different random seeds. 

\subsection{Hyperparameter selection}
The input window $k$ was set to $96$ hours using an informal hyperparameter search. The forecasting horizon $n$ was set to $12$ hours as this was a requirement.

Other hyperparameters of the neural networks were tuned using random search with 128 runs each. These hyperparameters are tuned on the global model version of each neural network and not re-tuned for other experiments. 

\subsection{Global vs local models}

To assess the effectiveness of global models, we evaluated two versions of each model. The first version is a global model that predicts the room temperature for all rooms using a single model trained with all the data. The second version is a local model where a separate model is trained for each room, denoted by the underscore \textit{p}. Additionally, we evaluated an alternative version of the transformer model that does not utilize a room embedding, denoted by the underscore \textit{ne}. 

We set the input window $k$ to 96 hours into the past, and the forecasting horizon $n$ to 12 hours into the future. Both the LSTM and Transformer models are residual models, and we followed Equation \ref{eq:residual} as this improved model performance. 

\subsection{Data Pre-processing}
To fairly compare global and local models, two different scaling strategies were used. An \textit{individual scaling strategy}, in which each of the 839 time series are individually scaled to be in the range $[0, 1]$ and a \textit{common scaling strategy} in which all air temperatures, including room temperature, outside temperature, common areas, among others, are all scaled together to be in the range $[0, 1]$ while all other time series are scaled individually.

\subsection{Evaluation Metrics}
The performance of the models was assessed using the mean average error (MAE) of the predicted room temperatures, this is  is defined in Equation \ref{eq:mae}.

\begin{equation}
MAE(Y, \hat{Y}) = \frac{1}{N} \sum_{n=1}^N |Y_n - \hat{Y}_n|
\label{eq:mae}
\end{equation}

Overall, the proposed models were evaluated and compared based on their ability to accurately predict the room temperature in industrial buildings.

Due to compute limitations each model was trained 8 times with different seeds, these results are reported in Table \ref{table:results}.

\section{Results and discussion}
A summary of the results of our experiments is reported in Table \ref{table:results}.

\begin{table}[ht!]

\begin{center}
\lineup
\begin{tabular}{lrr}
\br
             Model &     MAE &      Std \\
\mr
     $Persistence$ & 0.007400 & -- \\
          $LSTM_p$ & 0.007163 & 0.000140 \\
   $Transformer_p$ & 0.006995 & 0.000061 \\
     $Transformer$ & 0.004180 & 0.000052 \\
            $LSTM$ & 0.004161 & 0.000036 \\
$Transformer_{ne}$ & \textbf{0.004033} & 0.000027 \\
\br
\end{tabular}
\quad
\begin{tabular}{lrr}
\br
             Model &     MAE &      Std \\
\mr
     $Persistence$ & 0.049500 & -- \\
          $LSTM_p$ & 0.039025 & 0.000339 \\
   $Transformer_p$ & 0.037900 & 0.000203 \\
            $LSTM$ & 0.028783 & 0.000404 \\
$Transformer_{ne}$ & 0.027735 & 0.000223 \\
     $Transformer$ & \textbf{0.027012} & 0.000215 \\
\br
\end{tabular}
\end{center}
\caption{\label{table:results} Averaged results across 8 runs for each model for the common scaling (left) and individual scaling (right).} 
\end{table}

We find that the best Transformer model outperforms the best LSTM model for both common and individual scaling, with statistical significance ($p \textless 1.11e-06$ for common scaling, and $p \textless 2.98e-08$ for individual scaling). These results indicate that Transformer models are more effective than LSTM models for our task.

In terms of the impact of the room embedding, we find that for common scaling, including the room embedding significantly improves the performance of the global transformer model, with statistical significance ($p \textless 5.49e-06$). However, for individual scaling, the global transformer model without the room embedding performs significantly better ($p \textless 1.19-05$) than the model with the room embedding. 
The common scaling strategy simplifies batch inference by uniformly scaling all outputs, eliminating the need to track the room ID for each sample in a batch. This approach ensures consistent scaling across all outputs, facilitating the process of output interpretation and analysis.

Our results also show that the global models outperform the local models. Specifically, the global model lead to a performance increase of 16\% for the LSTM model and 40\% for the Transformer model for common scaling. For individual scaling, the global models perform 8\% better for the LSTM model and 27\% better for the Transformer model.

Finally, we find that the choice of scaling strategy depends on the specific model being used. For the Transformer model, individual scaling performs 4\% better than common scaling. For the $Transformer_{ne}$ model, common scaling provides a 1.5\% performance increase. In the case of the LSTM model, individual scaling performs the best, with a 1\% improvement over common scaling.

Overall, our results suggest that Transformer models are superior to LSTM models in terms of performance for our task, and that the choice of scaling strategy should be tailored to the specific model being used.
\section{Conclusion and future work}
This work presented a global Transformer architecture for indoor temperature forecasting in multi-room buildings, aiming to optimize energy consumption and reduce greenhouse gas emissions. The results demonstrated that Transformer models outperform LSTM models, with statistical significance, for both common and individual scaling approaches. The inclusion of room embedding significantly improves performance for common scaling, while the global Transformer model without room embedding performs better for individual scaling. Notably, the global models consistently outperformed the local models, offering the additional advantage of employing a single model for the entire building and resolving maintenance complexities. The choice of scaling strategy depends on the specific model used. Overall, the proposed Transformer architecture provides an efficient solution for accurate temperature forecasting, enabling energy optimization and emissions reduction in the building sector. Future research directions include analyzing the topology of the room embedding space, exploring the representation of the embedding, incorporating interpretability techniques, and leveraging pretrained Transformer models for generating synthetic data. These efforts will enhance the interpretability, performance, and robustness of the proposed approach for temperature forecasting in multi-room buildings.
\section*{References}
\bibliographystyle{plain}
\bibliography{references}

\begin{thebibliography}{10}

\bibitem{layernorm}
Jimmy~Lei Ba, Jamie~Ryan Kiros, and Geoffrey~E Hinton.
\newblock Layer normalization.
\newblock {\em arXiv preprint arXiv:1607.06450}, 2016.

\bibitem{IPCC_2022_WGIII_Ch_9}
L.~F. Cabeza, Q.~Bai, P.~Bertoldi, J.M. Kihila, A.F.P. Lucena, É. Mata,
  S.~Mirasgedis, A.~Novikova, and Y.~Saheb.
\newblock Buildings.
\newblock In P.R. Shukla, J.~Skea, R.~Slade, A.~Al Khourdajie, R.~van Diemen,
  D.~McCollum, M.~Pathak, S.~Some, P.~Vyas, R.~Fradera, M.~Belkacemi,
  A.~Hasija, G.~Lisboa, S.~Luz, and J.~Malley, editors, {\em Climate Change
  2022: Mitigation of Climate Change. Contribution of Working Group III to the
  Sixth Assessment Report of the Intergovernmental Panel on Climate Change},
  book section~9. Cambridge University Press, Cambridge, UK and New York, NY,
  USA, 2022.

\bibitem{glu}
Yann~N Dauphin, Angela Fan, Michael Auli, and David Grangier.
\newblock Language modeling with gated convolutional networks.
\newblock In {\em International conference on machine learning}, pages
  933--941. PMLR, 2017.

\bibitem{hochreiter1997long}
Sepp Hochreiter and J{\"u}rgen Schmidhuber.
\newblock Long short-term memory.
\newblock {\em Neural computation}, 9(8):1735--1780, 1997.

\bibitem{scalenorm}
Toan~Q Nguyen and Julian Salazar.
\newblock Transformers without tears: Improving the normalization of
  self-attention.
\newblock {\em arXiv preprint arXiv:1910.05895}, 2019.

\bibitem{glu_improve}
Noam Shazeer.
\newblock Glu variants improve transformer.
\newblock {\em arXiv preprint arXiv:2002.05202}, 2020.

\bibitem{su2021roformer}
Jianlin Su, Yu~Lu, Shengfeng Pan, Ahmed Murtadha, Bo~Wen, and Yunfeng Liu.
\newblock Roformer: Enhanced transformer with rotary position embedding.
\newblock {\em arXiv preprint arXiv:2104.09864}, 2021.

\bibitem{TIEN2022100198}
Paige~Wenbin Tien, Shuangyu Wei, Jo~Darkwa, Christopher Wood, and John~Kaiser
  Calautit.
\newblock Machine learning and deep learning methods for enhancing building
  energy efficiency and indoor environmental quality – a review.
\newblock {\em Energy and AI}, 10:100198, 2022.

\bibitem{Vaswani2017}
Ashish Vaswani, Noam Shazeer, Niki Parmar, Jakob Uszkoreit, Llion Jones,
  Aidan~N. Gomez, \L{}ukasz Kaiser, and Illia Polosukhin.
\newblock Attention is all you need.
\newblock In {\em Proceedings of the 31st International Conference on Neural
  Information Processing Systems}, NIPS'17, page 6000–6010, Red Hook, NY,
  USA, 2017. Curran Associates Inc.

\bibitem{vaswani2017attention}
Ashish Vaswani, Noam Shazeer, Niki Parmar, Jakob Uszkoreit, Llion Jones,
  Aidan~N Gomez, {\L}ukasz Kaiser, and Illia Polosukhin.
\newblock Attention is all you need.
\newblock {\em Advances in neural information processing systems}, 30, 2017.

\bibitem{prenorm}
Ruibin Xiong, Yunchang Yang, Di~He, Kai Zheng, Shuxin Zheng, Chen Xing,
  Huishuai Zhang, Yanyan Lan, Liwei Wang, and Tieyan Liu.
\newblock On layer normalization in the transformer architecture.
\newblock In {\em International Conference on Machine Learning}, pages
  10524--10533. PMLR, 2020.

\end{thebibliography}

\end{document}